\documentclass{article}
\pdfoutput=1

\usepackage{microtype}
\usepackage{graphicx}
\usepackage{subfigure}
\usepackage{booktabs} % for professional tables
\usepackage{multirow}
\usepackage[table,xcdraw]{xcolor}

\usepackage{hyperref}
\usepackage[accepted]{icml2019}

\icmltitlerunning{Probabilistic Discriminative Learning with Layered Graphical Models 
}

\usepackage{amsmath}
\usepackage{amsthm}
\usepackage{bm}
\usepackage{dsfont}
\usepackage{nth}
\usepackage{nicefrac}
\usepackage{titling}

\newcommand{\irange}[2]{\mathopen{[}#1\mathrel{{.}\,{.}}\nobreak#2\mathclose{]}}  % [a..b]

\DeclareMathOperator*{\argmin}{argmin}
\DeclareMathOperator*{\softmax}{softmax}
\DeclareMathOperator*{\flip}{flip}
\DeclareMathOperator*{\reshape}{reshape}
\DeclareMathOperator*{\sumsource}{sumSource}
\DeclareMathOperator*{\logsumexp}{logSumExp}
\DeclareMathOperator*{\relu}{ReLU}

\begin{document}

\twocolumn[
\icmltitle{Probabilistic Discriminative Learning with Layered Graphical Models 
}

\icmlsetsymbol{equal}{*}

\begin{icmlauthorlist}
\icmlauthor{Yuesong Shen}{tum}
\icmlauthor{Tao Wu}{tum}
\icmlauthor{Csaba Domokos}{bosch}
\icmlauthor{Daniel Cremers}{tum}
\end{icmlauthorlist}

\icmlaffiliation{tum}{Technical University of Munich, Munich, Germany}
\icmlaffiliation{bosch}{Bosch Center for Artificial intelligence, Renningen, Germany}

\icmlcorrespondingauthor{Yuesong Shen}{yuesong.shen @ tum.de}
\icmlcorrespondingauthor{Tao Wu}{tao.wu @ tum.de}
\icmlcorrespondingauthor{Csaba Domokos}{csaba.domokos @ de.bosch.com}
\icmlcorrespondingauthor{Daniel Cremers}{cremers @ tum.de}
%\icmlkeywords{Machine Learning, Graphical Model, Layered Graphical model, Variational Inference, Soft Clamping, ICML}

\vskip 0.3in
]

\printAffiliationsAndNotice{}  % leave blank if no need to mention equal contribution

\begin{abstract}
Probabilistic graphical models are traditionally known for their successes in generative modeling. 
In this work, we advocate layered graphical models (LGMs) for probabilistic discriminative learning. 
To this end, we design LGMs in close analogy to neural networks (NNs), that is, they have deep hierarchical structures and convolutional or local connections between layers. 
Equipped with tensorized truncated variational inference, our LGMs can be efficiently trained via backpropagation on mainstream deep learning frameworks such as PyTorch. 
To deal with continuous valued inputs, we use a simple yet effective soft-clamping strategy for efficient inference. 
Through extensive experiments on image classification over MNIST and FashionMNIST datasets, we demonstrate that LGMs are capable of achieving competitive results comparable to NNs of similar architectures, while preserving transparent probabilistic modeling. 
\end{abstract}

\section{Introduction}

Probabilistic graphical models \citep{pgmbook} offer an expressive approach to represent and reason with probability distributions. 
It has been successfully applied in various generative tasks such as extracting biologically similar visual features \citep{honglak08dbnv2} or inpainting occluded images \citep{mnih2011crbm, crbm_bp},
and is also widely used for structured prediction, such as semantic image segmentation \citep{densecrf, deeplab} in computer vision, or processing sequential data for domains like natural language processing \citep{manning1999nlp, collobert2011nlp} and signal processing \citep{chen03kalmanfilter}.

\subsection{Challenge}

For classical discriminative problems such as image classification, transparent probabilistic modeling would, to a large extent, facilitate model interpretability and uncertainty estimation \citep{Lipton18interp}. This goal, however, is challenging to achieve: State-of-the-art graphical model-based solutions are often limited to small scale or binary image datasets \citep{mnih2011crbm, crbm_bp}, due to intractable inference on general loopy graphs \citep{pgmbook}. Neural networks (NNs) \cite{dlbook}, though delivering currently the best performances \citep{krizhevsky12imagenet, resnet}, have no clear probabilistic interpretations, and are hard to analyze and vulnerable to human-indiscernible adversarial attacks \citep{goodfellow2015explaining}.

\begin{figure}[t]
\centering
\includegraphics[width=0.9\linewidth]{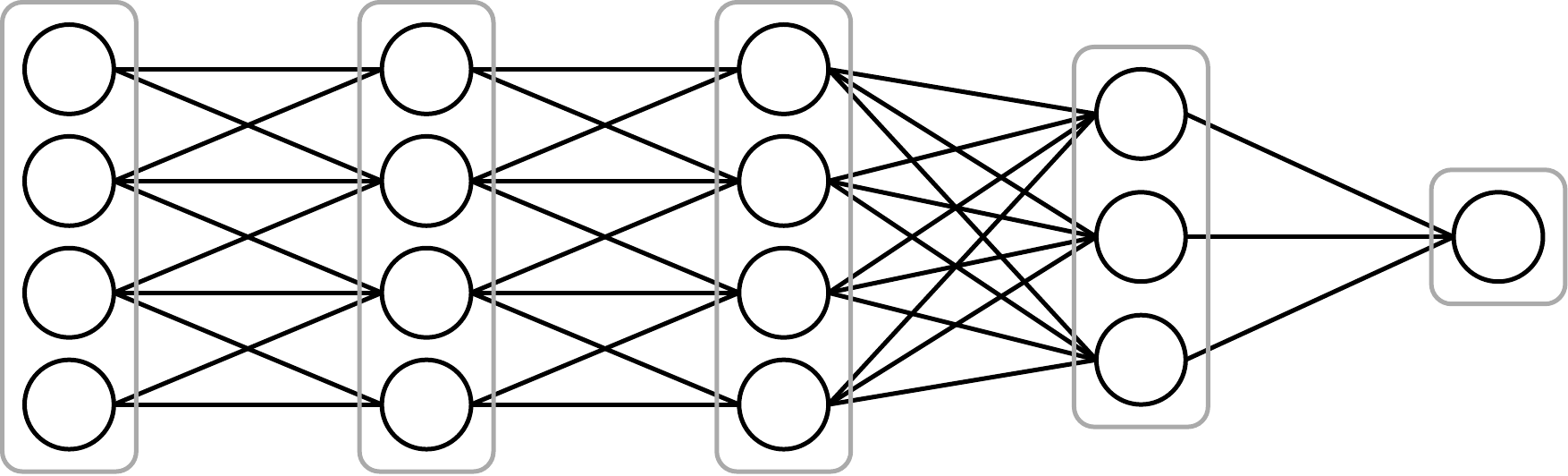}
\caption{Illustration of a layered graphical model consisting of five layers of nodes and four layerwise connections (two local and two dense connections). All connections are undirected.}
\label{fig:lgm_teaser}
\end{figure}

This paper addresses this challenge by proposing a layered graphical model framework, equipped with efficient inference and training, for probabilistic discriminative learning.

\subsection{Related work}

Conditional restricted Boltzmann machines (CRBMs) \citep{mnih2011crbm} extend restricted Boltzmann machines (RBMs) from generative to discriminative settings. Prior works \citep{mnih2011crbm, crbm_bp} have shown that by using approximative inference (via sampling or variational inference), CRBMs are able to handle binary image classification problems with noisy or occluded inputs. Especially, \citet{crbm_bp} demonstrates, with matricized operations, the effectiveness of loopy belief propagation \cite{bpbook}, which was deemed only practical for graphical models of moderate size \citep{mnih2011crbm}. However, the simplistic structure of CRBM limits its usage for more complex modeling. 

Using truncated Gibbs sampling, contrastive divergence (CD) \citep{Hin12} is designed to train RBMs efficiently. It can be further extended to train deep belief network \citep{dbn} and deep Boltzmann machine \citep{dbm} in a greedy, layerwise fashion. While these CD-trained models are effective for complex generative tasks \citep{dbn, dbm}, the probabilistic meaning of the whole model is somewhat lost due to the greedy layerwise approximation. Also, it is observed that CD is less effective in training conditional models \citep{mnih2011crbm, crbm_bp}. 

In terms of probabilistic modeling in deep learning,
variational autoencoder \citep{vae} models its latent space with mixture of Gaussians to generate data; Bayesian deep learning \citep{bnn-wu, mcdropout} introduces weights prior and applies Bayesian reasoning to model uncertainty \citep{kendall2017uncertainties}.
However, their probabilistic modelings are entangled with ``black-box'' NNs and the overall representations are not limpid.

As a side note, there also exist works that combine NNs and graphical models for structured prediction \citep{deeplab, crfasrnn, huang15lstmcrf}. 
In particular, \citet{crfasrnn} unrolls (truncated) mean-field inference and integrates it into the NN for end-to-end training. 

\subsection{Contributions}

The contributions of our paper are summarized as follows:

\begin{itemize}
\item We propose layered graphical models (LGMs) with hierarchical structures 
and convolutional and local connections in close analogy to convolutional NNs.
\item We integrate tensorized variational inference into MLE training of LGMs, and devise efficient training based on truncated inference and backpropagation.
To deal with continuous inputs such as grayscale images, 
we use a ``soft" clamping approach. 
\item Through extensive experiments we demonstrate
that LGMs can achieve competitive results comparable to NN baselines on various image classification tasks, 
while preserving transparent probabilistic modeling.
\end{itemize}

\section{Layered graphical models}
\label{sec:lgm}

To harness the power of hierarchical representation, in this work, we design a family of undirected graphical models with layered structure which we refer to as layered graphical models. The layered structure is attractive because:
1) It introduces a clear, compact and hierarchical representation of abstraction;
2) All associated computation can be performed in tensor form, which can be easily accelerated in modern computing systems. 

\subsection{General graphical representation}

The layered graphical model is a special instance of pairwise undirected graphical models where the nodes are arranged into layers. Denote the node and edge sets by $\mathcal{V}$ and $\mathcal{E}$, respectively. We model the joint distribution of random variables $\{x_i\}$, where $x_i \in \mathcal{X}_i$ % = \irange{1}{l_i}$ 
for each $i \in \mathcal{V}$, as Gibbs distribution given in energy form:
\begin{align}
P(\{x_i\}) = \frac{1}{Z} \exp \Big(-\sum_{i \in \mathcal{V}} E_i(x_i) - \sum_{(i, j) \in \mathcal{E}} E_{i,j}(x_i, x_j) \Big),
\label{eq:gibbs-distrib}
\end{align}
where $Z$ is the partition function, and $E_i$ and $E_{i,j}$ stand for the unary and pairwise energies, respectively.

Furthermore, we enforce the following constraints:

\begin{enumerate}
\item Each layer is homogeneous, i.e., with all its nodes having the same set of labels;
\item There are no intra-layer connections.
\end{enumerate}

Here we do not specify how the layers are connected between each other. We will discuss some possible connection types in Section~\ref{subsec:layercnnt}. Also, there is no constraint on the connection pattern of the hypergraph formed by layers: they can be connected a priori into loops, cliques, hypercubes, etc., although in this work we mainly focus on structures with chain-like connection patterns.

\subsection{Layerwise connections}
\label{subsec:layercnnt}

\begin{figure}[ht]
\centering
\includegraphics[width=0.45\linewidth, trim=0 -20 0 0]{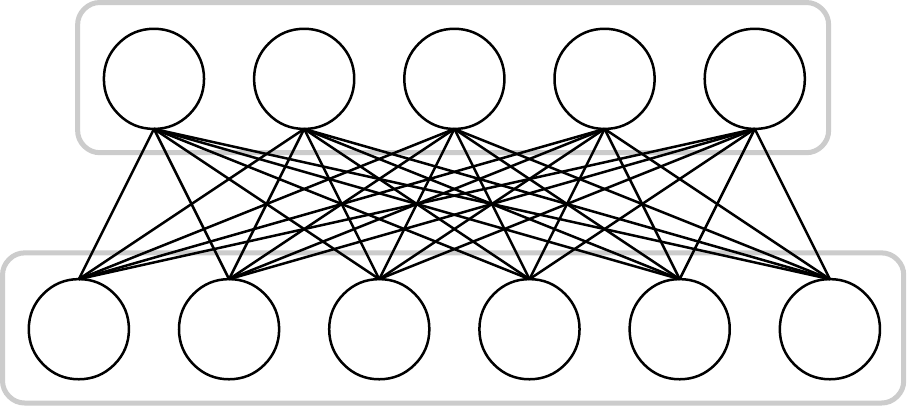}
\includegraphics[width=0.5\linewidth]{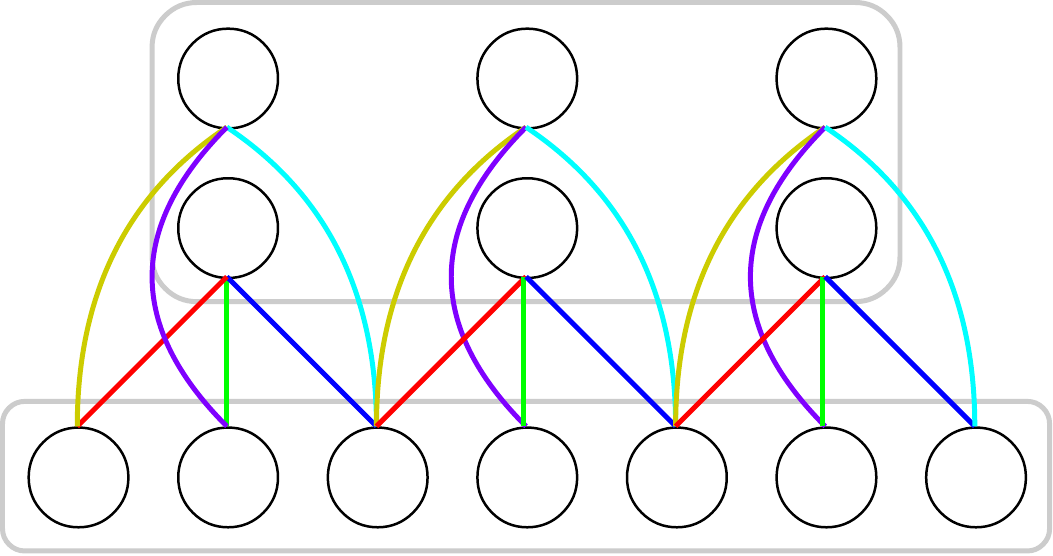}
\caption{Dense (left) and convolutional (right) connection. In the right subfigure, edges with the same color share the same weight. }
\label{fig:connections}
\end{figure}

One direct way to connect two layers is to connect all possible combinations and form a dense connection, as shown in Figure~\ref{fig:connections} on the left. This ensures that all possible interactions are considered by the learning process.

On the other hand, sometimes structured sparse connections are preferred for structured data such as images, where local patterns are predominant. In this case, one may wish to connect only local patches between layers in an LGM, which yields the convolutional connection (cf.\ Figure~\ref{fig:connections} right) if we enforce shared weights. We will also consider the variant without weight-sharing, which we refer to as local connection. Again, like in convolutional neural networks, we can customize parameters like kernel size, stride, dilation, etc.

\subsection{Connection to existing structures}

\paragraph{Boltzmann machine} RBM, conditional RBM and DBM are all special cases of LGM whose nodes have binary labels. Especially, the LBP updates in Section~\ref{subsec:infer} for LGM generalize the work of \citet{crbm_bp} and function on general multi-labeled layered structures. Compared to layered Boltzmann structures, higher cardinality nodes allow for natural representation of mutually exclusive cases, e.g.,\ output classes for classification and discretized input.

\paragraph{Neural network} LGMs and neural networks have similar connection patterns. However they represent different models in nature: neural networks are considered ``black-box'' universal function approximators, while LGMs offer transparent probabilistic modeling; accordingly, the inference process for neural networks is the feed-forward function evaluation, while LGMs require probabilistic or MAP inference procedures.

\section{Inference and learning}
\label{sec:inferlearn}

In this section, we present in details the efficient variational inference and learning methods for LGM. 

\subsection{Inference on LGM}
\label{subsec:infer}

\subsubsection{Variational inference methods}
\label{subsubsec:varinfer}

We begin with an overview of several variational inference methods, and customize them for LGM.

\paragraph{Mean field (MF)} The (na\"ive) mean field \citep{meanfield} approximates the joint distribution $P$ by a simpler distribution $Q$ consisting of the product of unary believes:
\begin{align}
Q(\{x_i\}) = \prod_{i \in \mathcal{V}} b_i(x_i).
\end{align}
By minimizing the KL-divergence $D(Q\|P)$, we obtain the following update formula:
\begin{align}
b_i = \softmax \Big( 
    - E_i - \sum_{\substack{j: (i,j) \in \mathcal{E} \\ 
                       x_j \in \mathcal{X}_j}} E_{i,j}(\cdot, x_j) \ b_j(x_j) \Big).
\label{eq:mf-ub-upd}
\end{align}

\paragraph{Loopy belief propagation (LBP)} The loopy belief propagation \citep{bpbook} generalizes the belief propagation from tree-structured graphs to general graphs. Its updates are expressed as follows:
\begin{align}
b_i &\propto \exp(- E_i) \ \prod_{j: (i,j) \in \mathcal{E}} m_{j \to i}, 
\label{eq:lbp-ub-upd} \\
%\end{align} 
%\begin{align}
m_{i \to j} &\propto \sum_{x_i \in \mathcal{X}_i} \left( \frac{\exp \big( -E_{i, j}(x_i, \cdot) \big)\ b_i(x_i)}{m_{j \to i}(x_i)}\right). 
\label{eq:lbp-m-upd} 
\end{align}
Here we directly construct the node-to-node messages since we have a pairwise model. As a side note, the above sum-product updates perform the probabilistic inference and can be easily modified to perform MAP inference (max-product) by changing the summation to maximization in Eq.~\eqref{eq:lbp-m-upd}.

\paragraph{Tree-reweighted message passing (TRW)} The tree-reweighted message passing  \citep{trw} approximates the original graph as a convex combination of its spanning trees. The update is similar to loopy belief propagation but involves defining a distribution over the set of spanning trees from which the edge appearance probabilities $\{\rho_{i,j}\}$ are deduced.

For LGMs having a tree structured layerwise connection, we can choose an arbitrary layer as root and construct spanning trees as combinations of mapping connections between connected layers following the leaf-to-root direction, provided that the mappings to the root layer are surjective. For classification tasks, the output layer typically has only one node, therefore the mappings onto it are trivially surjective.

Once $\{\rho_{i,j}\}$ are determined, the update formulas for TRW are obtained by adding a factor $\rho_{i,j}$ to each message $m_{j \to i}(x_i)$ in the belief update (Eq.~\eqref{eq:lbp-ub-upd}) and dividing the pairwise energy $E_{i,j}$ by $\rho_{i,j}$ in the message update (Eq.~\eqref{eq:lbp-m-upd}).

\subsubsection{Compact parametrization in log domain}

The variational inference methods in Section~\ref{subsubsec:varinfer} are all formulated in the domain of exponential of negative energy, for the ease of understanding. We found out that in practice it is beneficial to reformulate these inference updates in the log domain, since it allows for better numerical efficiency and stability in general.

Notably, the formulation in log domain allows us to easily remove redundant parameters and achieve compact representation: for unary believes and messages of $l$ labels, they can be represented as the softmax of $l-1$ parameters along with a fixed 0 last term; for unary and binary energies, we can reparametrize them so that one slice along each label direction can be zeroed-out. And it turns out that all inference updates we considered can be formulated directly with this compact representation. Further details are provided in the Supplementary Manuscript.

Also, we provide the tensorized implementations (including the pseudo-codes) of all aforementioned inference updates in the Supplementary Manuscript.

\subsection{Learning with LGM} \label{subsec:learning}

The parameters $\theta = \{E_i\} \cup \{E_{i,j}\}$ of an LGM are learned by maximizing the likelihood of the training data. Specifically, the variable nodes $\bm{x} = \{x_i\}$ of the LGM can be partitioned into three subsets: the input nodes $\bm{v}$, the hidden nodes $\bm{h}$ and the output nodes $\bm{y}$. For given data $\mathcal{D} = \{(\bm{v}_k^*, \bm{y}_k^*)\}_{k \in \mathcal{S}}$, we train $\theta$ to minimizes the negative log-likelihood (NLL):
\begin{equation}
\theta^* = \argmin_{\theta} - \sum_{k \in \mathcal{S}} \log P(\bm{y}=\bm{y}_k^* | \bm{v} = \bm{v}_k^*; \theta).
\end{equation}

Here the input nodes $\bm{v}$ are always observable and hence set as conditioned nodes, so that LGM does not need to model them. As we will see in Section~\ref{subsec:exp_partialinput}, we can easily extend the learning framework to the case where $\bm{v}$ is partially observable, and LGM will infer the missing part as a by-product of the learning process.

To find $\theta^*$ efficiently on a large dataset, we can estimate the likelihood using probabilistic inference, then perform mini-batch gradient descent with the negative log-likelihood loss. The mini-batch gradient can be computed using backpropagation, which is supported by mainstream deep learning frameworks. The probabilistic inference part, however, requires more attention, as it is intractable in general for loopy graphical models such as LGM. 

\subsubsection{Truncation of iterative inference}

We use the methods described in Section~\ref{subsubsec:varinfer} to perform approximative inference efficiently. They are all local and iterative updates, and we schedule them to run in parallel in the global scale, or layerwise if sequential updates are desired. Approximate inference such as LBP does not have convergence guarantee, but is observed to work well in practice \citep{loopybp}.

Furthermore, we truncate the inference procedure to a fixed number of iterations. The underlying rationale is that the truncated iterative inference provides a sufficiently good approximation of the prediction provided that the convergence takes place. In case of non-convergence, it instead provides a reasonable surrogate of the true prediction. The experiments in Section~\ref{subsubsec:expnbiter} identify with our reasoning.

As a remark, the truncated iterative inference was previously studied by \citet{truncatemp} using conditional random field on a small dataset. His observation agrees with ours, however without resorting to stochastic gradient and more complex models with hierarchical structures. 

\subsubsection{Training process of LGM}

Algorithm~\ref{algo:lgmtrain} summarizes the overall training process of LGM with $T$ inference iterations and backpropagation.

\begin{algorithm}[ht]
\caption{The training process of LGM}\label{algo:lgmtrain}
\begin{algorithmic}[1]
\STATE {\bfseries Input:} dataset $\mathcal{D} = \{(\bm{v}_k^*, \bm{y}_k^*)\}_{k \in \mathcal{S}}$, layered graphical model $\mathcal{G}$, number of inference steps $T$
\REPEAT
    \STATE $loss \gets 0$, draw mini-batch $\mathcal{B} \subset \mathcal{S}$
    \FORALL{$k\in\mathcal{B}$}
		\STATE Clamp input nodes of $\mathcal{G}$ to $\bm{v}_k^*$
		%\Comment{Initialization}
		\STATE Initialize $\mathcal{G}$ (w.r.t.~its believes, messages, etc.) 
		\FORALL{$t \in \irange{1}{T}$}
		%\Comment{Truncated inference}
		    \STATE Do parallel or sequential inference step
		\ENDFOR
		\STATE Get output probabilistic prediction $\bm{b}_k^{out}$
		%\Comment{Parameter update}
		\STATE $loss \gets loss + NLL(\bm{b}_k^{out}, \bm{y}_k^*)$
	\ENDFOR
	\STATE Update $\theta$ via backpropagation on $\nicefrac{1}{|\mathcal{B}|} \cdot loss$
\UNTIL{End of training}
\end{algorithmic}
\end{algorithm}

\subsubsection{Remarks on analytical gradient}

An alternative way of estimating the gradient of log-likelihood is to use its analytical expression: Let $E_c \in \theta$ be the energy depending on variables $\bm{x}_c$ of factor $c$ (either unary or pairwise in LGM), then the analytical gradient expression for $E_c$ can be written (with Iverson bracket) as:
\begin{align}
&\frac{\partial \log P(\bm{y}_k^* | \bm{v}_k^*; \theta)}{\partial E_c}(\hat{\bm{x}}_c) \notag\\
&= \mathds{E}_{P(\bm{h}|\bm{y}_k^*, \bm{v}_k^*)}\big[ [\bm{x}_c = \hat{\bm{x}}_c] \big] - \mathds{E}_{P(\bm{h}, \bm{y} | \bm{v}_k^*)}\big[ [\bm{x}_c = \hat{\bm{x}}_c] \big],
\label{eq:analytic_grad}
\end{align}
which takes the form of a difference of two expectations over an indicator function. Both expectations need to be estimated using probabilistic inference.

In principle, the analytical expression might offer better flexibility, since there is no need to build up the computation graph in the inference process as in the case of backpropagation. However, as shown in Section~\ref{subsubsec:exp_analyticgrad}, the analytical gradient (evaluated using truncated variational inference) yields significantly inferior results on loopy graphs, compared to the backpropagation approach. 
It is observed that the analytical gradient estimation is prone to inaccuracy from approximative inferences used for Eq.~\eqref{eq:analytic_grad}. In contrast, the backpropagation approach is always faithful to the truncated iterative inference even if it is inaccurate.

\section{Input modeling and soft clamping}
\label{sec:softclamp}

Continuous values such as grayscale intensity are commonly modeled in graphical models with their discretized representation. This is problematic since: 1) the high cardinality (e.g.,\ 256 for gray scale pixels) results in high computation cost and over-parametrization. 2) the natural ordering of the input is not preserved. Previous works on conditional RBM \citep{mnih2011crbm, crbm_bp} avoid this problem by using binarized inputs. While the first problem can be alleviated in some cases with a coarser quantization (see Section~\ref{subsec:requant}), a better input modeling is needed to properly tackle these problems.

In this section, we introduce soft clamping as a simple yet effective way to model continuous inputs, which enables efficient inference procedure.

\subsection{A soft clamped representation}

Soft clamping is based on a simple observation that any ranged continuous value (e.g.,\ grayscale intensity) can be regarded as a probabilistic state between its two endpoints (e.g.,\ black and white for image intensity). This allows us to model a continuous value using just one binary node. Accordingly, instead of considering an observation as a ``hard clamping'' of a node $V$ to a certain state, we ``soft clamp'' the mean $q^*$ of the Bernoulli distribution of the binary node to the observed value. 

\begin{figure}[ht]
\centering
\includegraphics[width=0.9\linewidth]{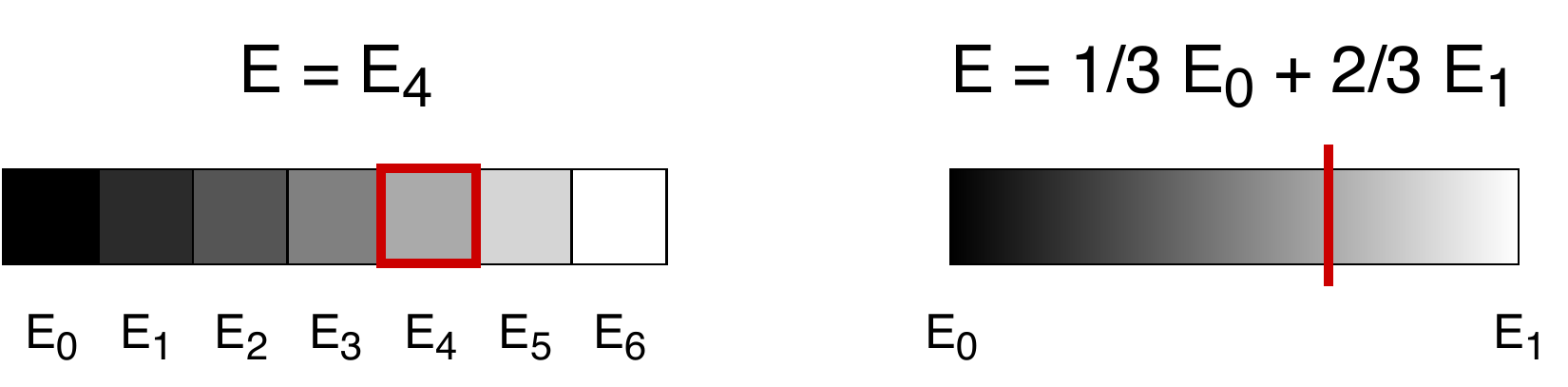}
\caption{A comparative illustration of discretized labeling (left) and soft clamping (right).}
\label{fig:softclamp}
\end{figure}

Thus using soft clamping, we could keep the original information and the ordering of the continuous observation, while reducing the label cardinality of the node to binary. Our objective function, however, is changed to minimizing the expected negative log-likelihood:
\begin{equation}
\theta^* = \argmin_{\theta}  \sum_{k \in \mathcal{S}} \mathds{E}_{\bm{V}_k}\big[ -\log \big( P(\bm{y}_k^*|\bm{V}_k; \theta) \big) \big].
\label{eq:softclampobj}
\end{equation}

\subsection{Approximation for efficient inference}

For each combination of hard clamped input $\{\bm{v}_k\}$, it is possible to estimate the negative log-likelihood $\sum_k  -\log (P(\bm{y}_k|\bm{v}_k; \theta))$ using an inference procedure. However, naively evaluating Eq.~\eqref{eq:softclampobj} would require that we perform an exponential number of inferences for all possible combination of $\bm{v}_k$, which is simply intractable.

Instead, we remark that:
\begin{align}
&\mathds{E}_{\bm{V}_k} \big[ -\log (P(\bm{y}_k^*|\bm{V}_k; \theta)) \big] \nonumber \\
&= \mathds{E}_{\bm{V}_k} \Big[ \log \Big( \sum_{\bm{h}, \bm{y}} \exp \big( -E(\bm{y}, \bm{h}|\bm{V}_k; \theta) \big) \Big) \nonumber \\
 & \qquad -\log \Big( \sum_{\bm{h}} \exp \big( -E(\bm{y}_k^*, \bm{h}|\bm{V}_k; \theta) \big) \Big) \Big] \label{eq:softclampexact}
\end{align}
which is an expectation of the difference of two convex ``log-sum-exp'' terms.

In soft clamping, we first compute the expected energy:
\begin{equation}
\bar{E}_k(\bm{y}, \bm{h}) = \mathds{E}_{\bm{V}_k}\big[ E(\bm{y}, \bm{h}|\bm{V}_k; \theta) \big]
\end{equation}
and then approximate Eq.~\eqref{eq:softclampexact} by:
\begin{align}
&\mathds{E}_{\bm{V}_k}\big[ -\log (P(\bm{y}_k^*|\bm{V}_k; \theta)) \big] 
\approx \log \big( \sum_{\bm{h}, \bm{y}} \exp(- \bar{E}_k (\bm{y}, \bm{h})) \big) \notag\\
& \qquad
 -\log \big( \sum_{\bm{h}} \exp(- \bar{E}_k (\bm{y}_k^*, \bm{h})) \big),
\label{eq:softclampapprox}
\end{align}
where both terms above refer to lower approximations, due to Jensen's inequality, of the respective terms in Eq.~\eqref{eq:softclampexact}.

It turns out that Eq.~\eqref{eq:softclampapprox} is actually the negative log-likelihood of the distribution with energy $\bar{E}_k$. It can then be computed efficiently with only one inference after computing $\bar{E}_k$, which in practice simply requires tensor product between the observations and binary energies of their connections (instead of slicing as in the hard clamping cases).

\subsection{Remarks on coarse quantization}
\label{subsec:requant}

A simple ``hard clamping'' alternative to reduce the complexity of the original discretization is to use a coarser quantization. This makes sense when the original precision is not critical for the task. As illustrated by the example of FashionMNIST below, the grayscale image classification problem fulfills this criteria:

\begin{figure}[ht]
\centering
\includegraphics[width=0.32\linewidth]{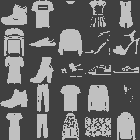}
\includegraphics[width=0.32\linewidth]{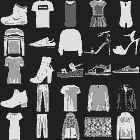}
\includegraphics[width=0.32\linewidth]{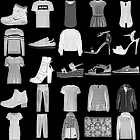}
\caption{(Re)quantization of FashionMNIST data samples to 2, 4 and 256 (original) colors.}
\label{fig:requant}
\end{figure}

In Figure~\ref{fig:requant}, we see that quantization to 4 colors can already give us sufficient information to recognize the shown objects. This indicates that there can be a trade-off between the input precision and the over-parametrization of learning model.

\section{Experiments}
\label{sec:experiments}

In this section, we present and analyze the experimental results for LGM on several image classification problems.

\subsection{General settings}
\label{subsec:expsettings}

We conducted a series of experiments on image classification using LGM to examine its properties. We tested over the MNIST~\citep{mnist} and the \mbox{FashionMNIST}~\citep{fashionmnist} datasets, both containing $28 \times 28$ grayscale images, 60,000 training samples, 10,000 test samples and 10 balanced classes. MNIST consists of black-and-white hand-written digits where grayscale is only used for anti-aliasing\footnote{See \url{http://yann.lecun.com/exdb/mnist/}, accessed on January \nth{12}, 2019.}, while \mbox{FashionMNIST} contains images from online shopping catalogs in true grayscale. For both datasets, we further split the training samples to 48,000 images (80\%) for training and 12,000 for validation (20\%).

In our implementation we use PyTorch \citep{pytorch} for GPU acceleration and auto-differentiation. For weight update we use Adam optimizer \citep{adam} with default settings. The batch size is set to 20 and trainings are stopped when the validation loss ceases to decrease. 

\subsection{Introspective tests}
\label{subsec:expdesignchoices}

We start with a series of experiments on MNIST to test out several aspects of LGM, namely the effect of truncated inference, soft clamping and the comparison with estimated analytical gradient learning method.

\subsubsection{Influence of truncation} 
\label{subsubsec:expnbiter}

\begin{figure}[ht]
\centering
\includegraphics[width=0.7\linewidth]{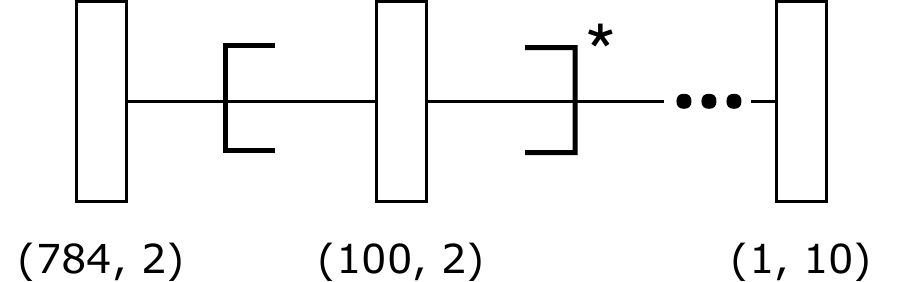}
\caption{Structure of sequential model with dense connections. The number of nodes and labels are indicated under each layer.}
\label{fig:truncinf}
\end{figure}

First of all, we study the behavior of truncated inference using a sequential LGM with dense connections and a certain number of binary hidden layers with 100 nodes in between, as shown in Figure~\ref{fig:truncinf}. The inputs are thresholded to binary.

Table~\ref{tab:seqstruct} summarizes the result of LGMs with varying depth and truncation. Each test was run with sequential LBP and same truncation was used for training and testing:

\begin{table}[ht]
\vskip -0.1in
\caption{Test accuracy of sequential models with $0$--$4$ hidden layers and $1$--$5$ inference iterations.}
\label{tab:seqstruct}
\vskip 0.1in
\centering
\begin{tabular}{llllll}
\toprule
  & 0 & 1 & 2 & 3 & 4 \\
\midrule
1 & 0.913 & \cellcolor[HTML]{CCCCCC}0.096 & \cellcolor[HTML]{CCCCCC}0.114 & \cellcolor[HTML]{CCCCCC}0.096 & \cellcolor[HTML]{CCCCCC}0.097 \\
2 & 0.911 & 0.947 & \cellcolor[HTML]{CCCCCC}0.114 & \cellcolor[HTML]{CCCCCC}0.114 & \cellcolor[HTML]{CCCCCC}0.096 \\
3 & 0.912 & 0.946 & 0.937 & \cellcolor[HTML]{CCCCCC}0.114 & \cellcolor[HTML]{CCCCCC}0.114 \\
4 & 0.911 & 0.941 & 0.946 & 0.937 & \cellcolor[HTML]{CCCCCC}0.098 \\
5 & 0.912 & 0.945 &  0.941 & 0.936 & 0.929 \\
\bottomrule
\end{tabular}
\end{table}

Here one can observe a clear separation of success and failure (in gray) cases, indicating that the number of inference iterations should be no smaller than the depth of the model (i.e.\ distance between the input and the output layer). If this is fulfilled, we can perceive that the truncation works quite well, and more iteration does not necessarily lead to better classification result.

Some further experiments indicate that using either longer or shorter iteration at test time seems to deteriorate the results in general. It is thus better to use exactly matched truncation during training and evaluation.

\subsubsection{Soft clamping for MNIST}
\label{subsubsec:expsoftclamp}

Also, we analyze the effect of using soft clamping for MNIST input. Table~\ref{tab:softclamp} provides the comparison results:

\begin{table}[ht]
\vskip -0.1in
\caption{Comparison of soft clamping against binary thresholding for MNIST: test accuracy of LGM with 1 hidden layer and 5 inference iterations.}
\label{tab:softclamp}
\vskip 0.1in
\centering
\begin{tabular}{cc}
\toprule
Soft clamp & Binary threshold\\ 
\midrule
 0.963 & 0.945 \\
\bottomrule
\end{tabular}
\end{table}

We see that even with the nearly binary MNIST dataset, soft clamping gives a visible boost to the result. Experiments in Section~\ref{subsec:exp_grayscale} further show the advantage of soft clamping for true grayscale data.

We will use soft clamping for LGMs in later experiments, unless stated otherwise.

\subsubsection{Analytic gradient} 
\label{subsubsec:exp_analyticgrad}

Additionally, we tested the approach of analytical gradient approximation. We compared 1) the case (Exact) of the sequential structure in Section~\ref{subsubsec:expnbiter} with no hidden layer so that the inference is exact; and 2) the case (Dense) with one hidden layer where the inference is approximative.

\begin{table}[ht]
\vskip -0.1in
\caption{Performance of analytic gradient in Exact and Dense cases (with number of inference iterations).}
\label{tab:cjinf}
\vskip 0.1in
\centering
\begin{tabular}{ccccc}
\toprule
     Exact & Dense(2) & Dense(5) & Dense(8) & Dense(20)\\
\midrule
0.910 & 0.770 & 0.874 & 0.879 & 0.880 \\
\bottomrule
\end{tabular}
\end{table}

The tests in Table~\ref{tab:cjinf} were run with sequential LBP and shows comparable result for exact inference. However, compared to the second column of Table~\ref{tab:seqstruct},we see that for approximative inference a significant performance drop can be observed, and performance deterioration can also be observed with other inference methods.

Also, we see a limited improvement for analytic gradient estimation with increased inference iterations, however it can not fully compensate the performance drop due to the inaccuracy from approximative inferences used for Eq.~\eqref{eq:analytic_grad}.

\subsection{Comparison of approximate inference methods}
\label{subsec:infer_compare}

\begin{figure}[ht]
\centering
\includegraphics[width=\linewidth , trim=0 25 0 15]{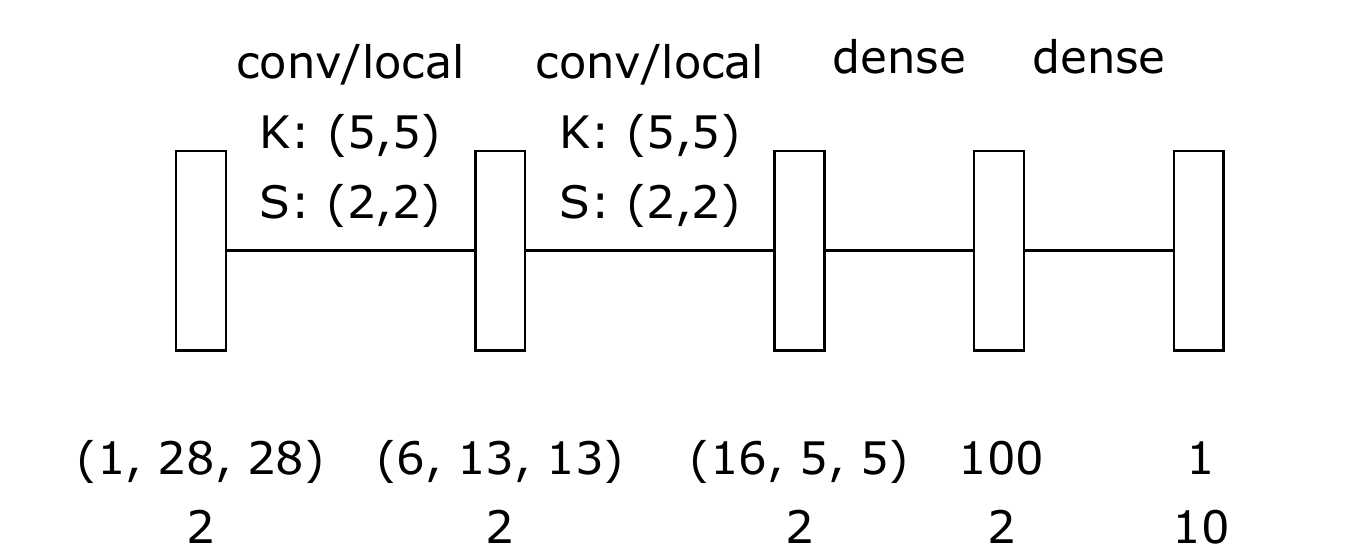}
\caption{Structure of LGM with two 2D conv/local connections of kernel size 5 and stride 2. Node shapes/sizes and label sizes are indicated for each layer. }
\label{fig:mininet}
\end{figure}

We then analyze the performance of the variational inference methods presented in Section~\ref{subsec:infer} (i.e.\ MF, LBP, TRW) with parallel and sequential scheduling (with prefix ``Par'' and ``Seq'', respectively), as well as the effect of different connections seen in Section~\ref{subsec:layercnnt}: we use the sequential model as in Section~\ref{subsubsec:expnbiter} with one hidden layer (Dense) for pure dense connection and the structure presented in Figure~\ref{fig:mininet} for convolutional (Conv) or local (Local) connections.

As baselines we used neural networks with sigmoid/rectified linear unit ($\relu$) activation function and similar architecture to the corresponding LGMs. We used 5 inference iterations for all LGM models.

For comparison, we also included the results for contrastive divergence (CD) with 1 and 10 Gibbs sampling steps. In this case, we pretrained all connections except the last one with greedy layerwise CD algorithm and then trained the last layer with softmax non-linearity as classifier. The results for local structure are not provided due to the lack of existing implementation for transposed local operation.

\begin{table}[ht]
\vskip -0.1in
\caption{Comparison of inference algorithms and neural network baselines with ``Dense'', ``Conv'' and ``Local'' structures.}
\label{tab:mpcompare}
\vskip 0.1in
\centering
\begin{tabular}{cccc}
\toprule
        & Dense & Conv & Local \\ 
\midrule
ParMF & 0.952 & 0.968 & 0.976\\
SeqMF & 0.951 & 0.927 & 0.936 \\
ParLBP & 0.962 & 0.968 & 0.974 \\
SeqLBP & 0.963 & 0.970 & 0.979 \\
ParTRW & 0.967 & 0.982 & \textbf{0.985} \\
SeqTRW & 0.966 & \textbf{0.983} & \textbf{0.984} \\
CD1 & 0.938 & 0.939 & -- \\
CD10 & 0.936 & 0.935 & -- \\
NN($\sigma$) & 0.970 & \textbf{0.984} & 0.983 \\
NN(relu) & \textbf{0.973} & 0.982 & \textbf{0.984} \\
\bottomrule
\end{tabular}
\end{table}

Table~\ref{tab:mpcompare} reports the comparison. We observe that
\begin{itemize}
\item Overall, convolutional and local connections result in better performance, both for neural networks and LGMs;
\item The variational inference approaches clearly out-perform layerwise contrastive divergence on all architectures, while tree-reweighted message passing yields the best results for variational inference;
\item Compared to neural networks, LGMs achieve comparable results with convolutional/local connections but worse results with only dense connections.
\end{itemize}

\subsection{Dealing with grayscale images}
\label{subsec:exp_grayscale}

Furthermore, we test our system for grayscale image classification using FashionMNIST.

\subsubsection{Soft clamping v.s. requantization}

Recall from Section~\ref{sec:softclamp} that we have discussed two methods for efficiently dealing with grayscale images: soft clamping and coarse quantization. We will validate our intuition that a coarse quantization is sufficient for the classification task, as well as compare the performance of these two approaches.

An additional issue for requantization is to extend the hidden part to accommodate the increase of input label space. To achieve this, we can increase either the number of nodes for hidden layers, or the number of labels. 
To compare these two approaches, we used a dense sequential model with one hidden layer. For input with $2^n$ colors:
\begin{itemize}
\item The ``N'' approach extends the hidden layer size to $100 \times n$ while keeping the binary labeling;
\item The ``L'' approach extends the label size to $n+1$ while keeping the layer size to 100.
\end{itemize}

Table~\ref{tab:fashionmnist} shows the results with sequential LBP. We reduced the batchsize to 4 for the tests with 32 colors and to 1 for 256 colors in order to limit the runtime memory load. We also consider the no-scaling baseline (``F'').

\begin{table}[ht]
\vskip -0.1in
\caption{Comparison of soft clamping and quantization to 2, 4, 8, 32, 256 colors with different scaling strategies: for $2^n$ colors, ``F'' fixes the hidden layer to be binary of size 100, ``N'' extends its size to $100 \times n$, and ``L'' extends the cardinality to $n+1$.} \label{tab:fashionmnist}
\vskip 0.1in
\centering
\begin{tabular}{ccccccc}
\toprule
 & 2 & 4 & 8  & 32 & 256 & Soft\\
\midrule
F & 0.801 & 0.813 & 0.804  & 0.794 & 0.775 & \textbf{0.865} \\
N & N/A & 0.816 & 0.800  & 0.790 & 0.734 & N/A \\
L & N/A & 0.817 & 0.812  & 0.777 & 0.725 & N/A \\
\bottomrule
\end{tabular}
\end{table}

We conclude from Table~\ref{tab:fashionmnist} that finer quantization does not necessarily improve the result for LGMs. Instead, it seems to cause over-fitting. Also, soft clamping approach out-performs requantization by a considerable margin.

\subsubsection{Results on FashionMNIST}

Considering the previous results, we now perform a comparison of the two approaches using convolutional/local connections with neural network baselines. For soft clamping and neural network baselines, we reuse the structure shown in Figure~\ref{fig:mininet}, while for requantization we set the input to have 4 colors, and extend the first hidden layer to 3 labels. Sequential TRW is used for all LGM structures.

\begin{table}[ht]
\vskip -0.1in
\caption{Comparison between message passing algorithms and neural network baselines.}
\label{tab:fashionmnist_nncomp}
\vskip 0.1in
\centering
\begin{tabular}{ccccc}
\toprule
       & Soft & Requant. & NN($\sigma$) & NN(relu) \\
\midrule
Conv  & 0.884 & 0.865 & 0.889 & \textbf{0.894} \\
Local & \textbf{0.894} & 0.863 & \textbf{0.894} & \textbf{0.894} \\
\bottomrule
\end{tabular}
\end{table}

Table~\ref{tab:fashionmnist_nncomp} lists the results. Again, the soft clamping approach outperforms the requantization approach. And we see that 
with soft clamping LGM is able to attain accuracy comparable to neural networks with similar architectures. 

\subsection{Classification of partially observable inputs}
\label{subsec:exp_partialinput}

Finally, we experiment on MNIST and FashionMNIST with partially observable inputs: in this case, each input pixel has a certain probability $p_{\text{obs}}$ to be observed. To handle this, LGM models the input explicitly by taking the unobserved pixels as unclamped nodes in the inference process. For the neural network baselines, we heuristically fill up the missing pixels with gray value 0.5. 

Also, to account for the uncertainty of the output caused by unreliable input, we apply smoothing to the ground-truth labels: instead of the one-hot representation, we set the probability of correct label to $1 - \nicefrac{9 \varepsilon}{10}$ and $\nicefrac{\varepsilon}{10}$ for the rest. We fix 
$\varepsilon=0.1$ and observe consistent improvement over all methods in our experiments.

\begin{table}[ht]
\vskip -0.1in
\caption{Accuracies on MNIST with partially observed input (0.3 and 0.7 visible) using LGM and neural network baselines.}
\label{tab:exp_partial_input_m}
\vskip 0.1in
\centering
\begin{tabular}{cccc}
\toprule
       & LGM & NN($\sigma$) & NN(relu) \\
\midrule
 Conv (0.3)  & 0.952 & 0.954 & \textbf{0.956} \\
 Local (0.3) & \textbf{0.956} & 0.951 & 0.952 \\
 Conv (0.7)  & 0.978 & 0.984 & \textbf{0.986} \\
 Local (0.7) & 0.984 & 0.983 & \textbf{0.985} \\
\bottomrule
\end{tabular}
\end{table}

\begin{table}[ht]
\vskip -0.1in
\caption{Accuracies on FashionMNIST with partially observed input (0.3 and 0.7 visible) using LGM and neural network baselines.}
\label{tab:exp_partial_input_fm}
\vskip 0.1in
\centering
\begin{tabular}{cccc}
\toprule
       & LGM & NN($\sigma$) & NN(relu) \\
\midrule
Conv (0.3)  & 0.828 & 0.822 & 0.830 \\
Local (0.3) & 0.830 & 0.822 & \textbf{0.832} \\
Conv (0.7)  & 0.862 & 0.866 & 0.878 \\
Local (0.7) & \textbf{0.882} & 0.879 & \textbf{0.881} \\
\bottomrule
\end{tabular}
\end{table}

Tables~\ref{tab:exp_partial_input_m} and \ref{tab:exp_partial_input_fm} summarize the results for experiments with partial input. Compared to neural network baselines, LGM yields slightly sub-optimal performance in some cases, however the results are still comparable in general.

\begin{figure}[ht]
\centering
\includegraphics[scale=0.5]{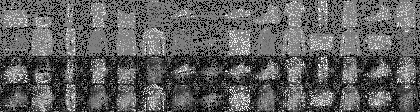}
\caption{Samples from Local LGM for FashionMNIST with 30\% visible input. The silhouettes become clearer with filled believes.}
\label{fig:recon_fm}
\vspace{-0.1in}
\end{figure}

\begin{figure}[!ht]
\begin{minipage}{.2\linewidth}
\centering
\begin{tabular}{c}
\includegraphics{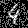}  \\
\includegraphics{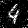}
\end{tabular}
\end{minipage}
\begin{minipage}{.79\linewidth}
\centering
\includegraphics[scale=0.28]{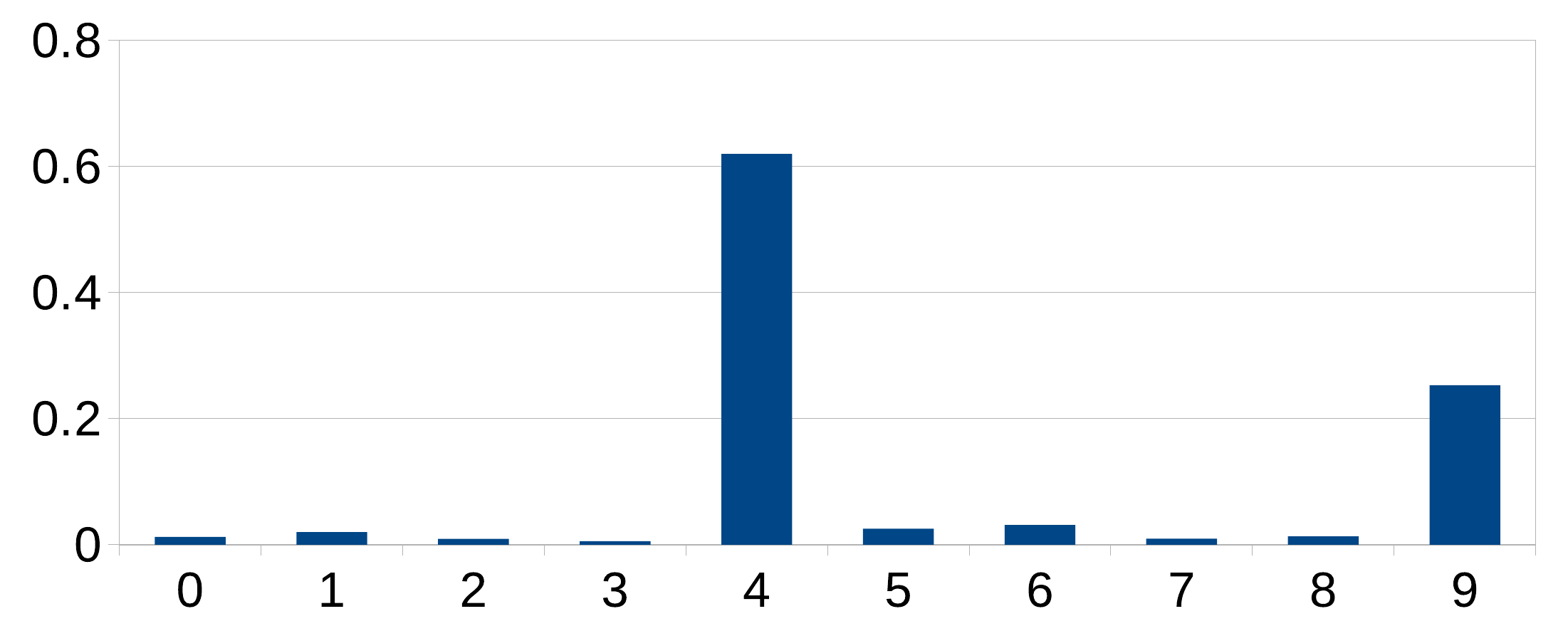}
\end{minipage}
\caption{An ambiguous sample from MNIST with 70\% visible input, its  belief-filling with Local LGM, and its probabilistic prediction: the model shows uncertainty between 4 and 9.}
\label{fig:recon_m}
\end{figure}

Interestingly, the probabilistic modeling of LGM provides additional insights. 
For example,
we obtain the believes of the missing input pixels as an outcome of probabilistic inference; see visualizations in Figures~\ref{fig:recon_fm} and \ref{fig:recon_m}. Figure~\ref{fig:recon_m} also shows that the end-to-end probabilistic modeling of LGM is able to correctly handle ambiguous inputs.

\section{Conclusion}

We propose the layered graphical model framework for efficient probabilistic discriminative learning. Combining 
\begin{itemize}
\item
layered architecture,
\item
local or convolutional connections, 
\item
truncated variational inference with backpropagation,
\item
soft clamping, 
\end{itemize}
our layered graphical models are able to go beyond existing application range of probabilistic graphical models. 
As shown by Sections~\ref{subsec:infer_compare} and \ref{subsec:exp_grayscale}, they achieve comparable performances vis-a-vis resembling neural networks on grayscale image classification.

Compared to neural networks, layered graphical models additionally provide a transparent probabilistic representation, which, as indicated by Section~\ref{subsec:exp_partialinput}, allows for natural modeling of uncertain inputs and inference of missing information. 

We expect our work to open new opportunities for graphical models in uncertainty modeling and interpretable learning. 

\clearpage
\bibliography{reference}
\bibliographystyle{icml2019}

\clearpage
\twocolumn[
\icmltitle{Supplementary Manuscript: \\
Probabilistic Discriminative Learning with Layered Graphical Models 
}
\vskip 0.3in
]
\appendix

Here we provide additional details for the practical implementation of LGM. The demo code is provided at \url{https://github.com/tum-vision/lgm}.

\section{Parametrization of LGM}

In this section, we specify the way layers and connections are parametrized in our implementation.

\subsection{Compact parametrization in log domain}
\label{subsec:minimalrepr}

To represent the parameters and the intermediate states of LGM in a compact and non-redundant way, we make use of the following compact representation in logarithmic domain ($\mathcal{X}_i = \irange{1}{l_i}$ denotes the label set of node $i$):

\paragraph{Energies} The unary and pairwise energies are the parameters of LGM which determine the joint distribution. The energy terms are defined up to a constant, and we can re-parametrize them to satisfy the following constraints:
\begin{align}
\forall i \in \mathcal{V} &: E_{i}(1) = 0; \\
\forall (i, j) \in \mathcal{E}, \forall (x_i, x_j) &: E_{i,j}(1, x_j) = E_{i,j}(x_i, 1) = 0.
\end{align}

\paragraph{Believes, messages} The inference methods that we consider in this work are all iterative and the intermediate states such as unary believes and (normalized) messages are all probability-like. For example, the unary belief $b_i$ has the following constraints:
\begin{align}
\forall x_i \in \mathcal{X}_i: b_i(x_i) \geq 0 \text{ and } \sum_{x_i \in \mathcal{X}_i} b_i(x_i) = 1.
\end{align}
We can thus define the following log-form:
\begin{align}
\forall  x_i \in \mathcal{X}_i: \tilde{b}_i(x_i) = \log(b_i(x_i)) - \log(b_i(1)),
\end{align}
so that we have
\begin{align}
\tilde{b}_i (1) &= 0, \\
\softmax(\tilde{b}_i) &= b_i.
\end{align}
The log-domain believes $\{\tilde{b}_i(x_i): x_i \in \mathcal{X}_i \setminus \{1\}\}$ are thus unconstrained and match the intrinsic degrees of freedom in the original parameterization $\{b_i(x_i): x_i \in \mathcal{X}_i\}$. For normalized message $m_{i \to j}$, we define $\tilde{m}_{i \to j}$ similarly.

Note that the choice of always ``zeroing-out'' the first label is merely for notational convenience, the label choice can be arbitrary for each node.

\subsection{Parametrization for dense connection} 
For layers $p, q$ having $n^p, n^q$ nodes with $l^p, l^q$ labels respectively, the full (dense) connection will introduce $n^p \times n^q$ edges. Using the minimal representation described in Section~\ref{subsec:minimalrepr}, we can represent the unary energies in e.g., layer $p$ using a tensor $\bm{V}^p$ of shape $(n^p, l^p-1)$, and the pairwise energies between $p,q$ with a tensor $\bm{W}^{p,q}$ of shape $(n^p, n^q, l^p-1, l^q-1)$.

\subsection{Parametrization for local/convolutional connection} 

Similar to a $d$-dimensional convolutional layer in a neural network, using the minimal representation, each layer $p$ will have $c^p$ channels of nodes arranged in shape $\bm{s}^p = (s_1^p, \dots, s_d^p)$ so that the unary energy tensor $\bm{V}^p$ is of shape $(c^p, \bm{s}^p, l^p-1)$, and we define the kernel size (i.e.\ patch size) between $p, q$ as $\bm{k}^{p,q} = (k_1^{p,q}, \dots, k_d^{p,q})$ so that the pairwise energy tensor $\bm{W}^{p,q}$ has shape $(c^p, c^q, \bm{k}^{p,q}, \bm{s}^q, l^p-1, l^q-1)$ for local connection. The convolutional connection refers to local connection with shared energies across all patches, hence yielding $\bm{W}^{p,q}$ of shape $(c^p, c^q, \bm{k}^{p,q}, l^p-1, l^q-1)$.

\subsection{Equivalence between \texorpdfstring{$\bm{W}^{p,q}$}{W_pq} and \texorpdfstring{$\bm{W}^{q,p}$}{W_qp}} \label{subsubsec:equirepr}

With the above tensor representation, $\bm{W}^{p,q}$ and $\bm{W}^{q,p}$ typically have different shapes. Nevertheless, they represent the same set of parameters arranged in different orders, and we can define a ``$\flip$'' operation to transform between these two shapes. 

\section{Efficient variational inference}
\label{sec:vi_implementation}

In Section~\ref{subsec:infer}, we reviewed several variational inference methods and their iterative updates. Here we provide explicitly the updates with compact parametrization.

\subsection{Inference updates with compact parametrization}
\label{subsec:vi_compactparam}

\paragraph{Mean field (MF)} For mean field we have
\begin{align}
\tilde{b}_i = &-E_i - \sum_{\substack{j: (i,j) \in \mathcal{E} \\ 
                                      x_j \in \mathcal{X}_j}} E_{i,j}(\cdot, x_j)  \ \softmax (\tilde{b}_j)(x_j).
\label{eq:mf-ub-upd-mr}
\end{align}

\paragraph{Loopy belief propagation (LBP)} The loopy belief propagation updates become the following:
\begin{align}
\tilde{b}_i(x_i) = - E_i (x_i) + \sum_{j: (i,j) \in \mathcal{E}} \tilde{m}_{j \to i}(x_i),
\label{eq:lbp-ub-upd-mr}
\end{align}
\begin{gather}
\tilde{m}_{i \to j} = \log \sum_{x_i \in \mathcal{X}_i} \exp ( -E_{i, j}(x_i, \cdot) + \tilde{b}_i(x_i) - \tilde{m}_{j \to i}(x_i) ) \nonumber \\
- \log\sum_{x_i \in \mathcal{X}_i}\exp ( \tilde{b}_i(x_i) - \tilde{m}_{j\to i}(x_i) ).
\label{eq:lbp-m-upd-mr}
\end{gather}

\paragraph{Tree-reweighted message passing (TRW)} With $\{\rho_{i,j}\}$ as defined in Section~\ref{subsubsec:varinfer}, the update for tree-reweighted message passing becomes:
\begin{align}
\tilde{b}_i(x_i) = - E_i (x_i) + \sum_{j: (i,j) \in \mathcal{E}} \rho_{i,j} \tilde{m}_{j \to i}(x_i),
\label{eq:trw-ub-upd-mr}
\end{align}
\begin{gather}
\tilde{m}_{i \to j} = \log \sum_{x_i \in \mathcal{X}_i} \exp \big( -\frac{E_{i, j}(x_i, \cdot)}{\rho_{i,j}} + \tilde{b}_i(x_i) - \tilde{m}_{j \to i}(x_i) \big) \nonumber \\
- \log\sum_{x_i \in \mathcal{X}_i}\exp \big( \tilde{b}_i(x_i) - \tilde{m}_{j\to i}(x_i) \big).
\label{eq:trw-m-upd-mr}
\end{gather}
We see that they are quite similar to Eq. \eqref{eq:lbp-ub-upd-mr} and \eqref{eq:lbp-m-upd-mr}. We will thus omit its pseudo-code in Section~\ref{subsec:vi_implement}

\subsection{Implementation of efficient updates for LGM}
\label{subsec:vi_implement}

Based on the Eq.~\eqref{eq:mf-ub-upd-mr},\eqref{eq:lbp-ub-upd-mr},\eqref{eq:lbp-m-upd-mr}, we implement inference updates in tensor form using the minimal representation. 
Denote by $\bm{B}^{p}$ the unary belief tensor for layer $p$ and by $\bm{M}^{p \to q}$ the incoming message tensor from $p$ to $q$. Both $\bm{B}^{p}$ and $\bm{M}^{p \to q}$ take (log-form) minimal representation, and $\bm{B}^{p}$ is of the same shape as $\bm{V}^{p}$ while $\bm{M}^{p \to q}$ has a similar shape than $\bm{W}^{p,q}$, except that it does not have the label dimension of layer $p$, nor does it have shared weights. We also define the following tensor operations:

\begin{itemize}
\item $\reshape$ adds necessary broadcastable dimensions to inputs so that the corresponding dimensions are aligned for elementwise operations;
\item $\flip$ transforms the input to its alternative representation, as described in Section~\ref{subsubsec:equirepr};
\item $\sumsource$ sums over the dimensions related to the source layer.
\end{itemize}

\subsubsection{Mean field}

The inference update step for mean field with LGM are shown in Algorithm~\ref{algo:mf}. We use ``$\odot$'' for the elementwise product between tensors, ``$\softmax^*$'' for the $\softmax$ operation over the label dimension which takes into account the implicit label, and ``$\bm{C}^{q, p}$'' to denote the contribution to $\bm{B}^{p}$ update from layer $q$. 

\begin{algorithm}[ht]
\caption{Belief update for mean field}\label{algo:mf}
\begin{algorithmic}[1]
\STATE \textbf{Inputs:} $\{(\bm{W}^{l,p},\bm{B}^{l}): (l, p)\in \mathcal{E}\}, \bm{V}^{p}$
\STATE \textbf{Output:} $\bm{B}^{p}$
	\STATE $\bm{B}^{p} \gets -\bm{V}^{p}$
	\FORALL{$(\bm{W}^{q, p},\bm{B}^{q}) : (q, p)\in \mathcal{E}$}
	    \STATE $\bm{\tilde{W}}^{q, p},\bm{\tilde{B}}^{q} \gets \reshape(\bm{W}^{q, p},\bm{B}^{q})$
	    \STATE $\bm{C}^{q, p} \gets -\bm{\tilde{W}}^{q, p} \odot \softmax^*(\bm{\tilde{B}}^{q})$
		\STATE $\bm{B}^{p} \gets \bm{B}^{p} +\sumsource (\bm{C}^{q, p})$
	\ENDFOR
\end{algorithmic}
\end{algorithm}

\subsubsection{Loopy belief propagation (LBP)}

Algorithms~\ref{algo:lbp_m} and \ref{algo:lbp_b} show the inference update steps for loopy belief propagation on LGM. Here ``$\logsumexp^*$'' denotes the $\logsumexp$ operator over the latter label dimension while taking into account the implicit label. 
Without showing further details, we remark that LBP can be turned into MAP inference by replacing $\logsumexp^*$ with $\relu\circ\max$ in Algorithm~\ref{algo:lbp_m}. With minor changes as discussed in Section~\ref{subsubsec:varinfer}, LBP can also be adapted for tree-reweighted message passing.

\begin{algorithm}[ht]
\caption{Message update for LBP}\label{algo:lbp_m}
\begin{algorithmic}[1]
\STATE \textbf{Inputs:} $\bm{M}^{q \to p}, \bm{B}^{p}, \bm{W}^{q, p}$
\STATE \textbf{Output:} $\bm{M}^{p \to q}$
\STATE $\bm{\tilde{M}}^{q \to p}, \bm{\tilde{B}}^{p},\bm{\tilde{W}}^{q, p} \gets \reshape (\bm{M}^{q \to p}, \bm{B}^{p}, \bm{W}^{q, p})$
\STATE $\bm{\bar{M}}^{p \to q} \gets - \logsumexp^*(\bm{\tilde{B}}^{p} - \bm{\tilde{M}}^{q \to p}) + \logsumexp^*(- \bm{\tilde{W}}^{q, p} + \bm{\tilde{B}}^{p} - \bm{\tilde{M}}^{q \to p}) $
\STATE $\bm{M}^{p \to q} = \flip (\bm{\bar{M}}^{p \to q})$
\end{algorithmic}
\end{algorithm}

\begin{algorithm}[ht]
\caption{Belief update for LBP}\label{algo:lbp_b}
\begin{algorithmic}[1]
\STATE \textbf{Inputs:} $\{\bm{M}^{l \to p}: (l, p)\in \mathcal{E}\}, \bm{V}^{p}$
\STATE \textbf{Output:} $\bm{B}^{p}$
\STATE $\bm{B}^{p} \gets -\bm{V}^{p}$
\FORALL{$\bm{M}^{q \to p} : (q, p)\in \mathcal{E}$}
	\STATE $\bm{B}^{p} \gets \bm{B}^{p} + \sumsource(\bm{M}^{q \to p})$
\ENDFOR
\end{algorithmic}
\end{algorithm}

\end{document}